# Attention-based Multi-instance Neural Network for Medical Diagnosis from Incomplete and Low Quality Data

Zeyuan Wang[1,3], Josiah Poon[1], Shiding Sun[2], Simon Poon[1*]
[1]School of Computer Science, The University of Sydney, Syndey, Australia
[2]School of Mathematics, Renmin University of China, Beijing, China
[3]Beijing Medicinovo Technology Co.,Ltd., Beijing, China
[1,3]zwan7221@uni.sydney.edu.au, [1]{josiah.poon, simon.poon}@sydney.edu.au, [2]sunshiding@ruc.edu.cn

*Abstract*— One way to extract patterns from clinical records is to consider each patient record as a bag with various number of instances in the form of symptoms. Medical diagnosis is to discover informative ones first and then map them to one or more diseases. In many cases, patients are represented as vectors in some feature space and a classifier is applied after to generate diagnosis results. However, in many real-world cases, data is often of low-quality due to a variety of reasons, such as data consistency, integrity, completeness, accuracy, etc. In this paper, we propose a novel approach, attention based multi-instance neural network (AMI-Net), to make the single disease classification only based on the existing and valid information in the real-world outpatient records. In the context of a patient, it takes a bag of instances as input and output the bag label directly in end-to-end way. Embedding layer is adopted at the beginning, mapping instances into an embedding space which represents the individual patient condition. The correlations among instances and their importance for the final classification are captured by multi-head attention transformer, instance-level multi-instance pooling and bag-level multi-instance pooling. The proposed approach was test on two non-standardized and highly imbalanced datasets, one in the Traditional Chinese Medicine (TCM) domain and the other in the Western Medicine (WM) domain. Our preliminary results show that the proposed approach outperforms all baselines results by a significant margin.

*Keywords—medical diagnosis, low-quality data, multi-instance learning, attention mechanism, deep learning*

I. INTRODUCTION

In many real-world observational studies, data is collected from various real-life applications instead of controlled experiment settings. These observational data are subject to data quality concerns such as: (i) data accuracy, (ii) data completeness, (iii) data consistency, and (iv) data balance [1]. More importantly, in real life, clinical decisions are made only from a few informative and valuable attributes, i.e., features, instead of the entire patient record. The use of computational models to find key information from a large amount of incomplete and low-quality data to generate the solid diagnosis results, has become a topic of broad interest.

From the machine learning perspective, this scenario is known as weakly supervised learning (WSL) and multi-instance learning (MIL) is a typical one [2]. This perspective considers the input sample as a bag of instances and only the bag label is given. Through learning and training, MIL models allow to predict the labels of new bags with containing instances. MIL was first proposed by Dietterich et. al [3] for drug molecule activity prediction, and has been widely applied in many fields, including medical imaging and video analysis [4, 5], syndrome differentiation in Tradition Chinese Medicine (TCM) [6, 7], pulmonary embolism and colon cancer detection [8] and retinal nerve fiber layer visibility classification [9]. In this paper, we mainly emphasize the application of MIL for the single disease diagnosis, i.e., binary classification on a single task.

Based on the definition of MIL, the bag is labeled positive only if at least one instance is positive, otherwise, the bag is labeled negative. In the context, capturing correlations among instances and finding the most informative instances play major roles. In many previous studies, the focus was on the latter, i.e., important instances detection, such as EM-DD [10], mi-SVM [11], mi-Graph [12] and miFV [13]. It is important to note that in some circumstances, if the correlations among instances are neglected, the prediction might be misled. For example, the bag is labeled as beach, only if the sky, ocean and sand co-occur [14]. Considering a complex progress like medical diagnosis, not only clinicians need to assess risk factors independently, but also take the influence of their co-occurrence into account. This is one of our starting points that we aim to measure instance correlations in the model building process.

Multi-instance neural network was first proposed by Ramon et al. [15]. This approach computes instance probabilities to be further processed by the log-sum-exp operator to get the bag probability and the whole process is trained in end-to-end way. Their work demonstrates the effectiveness and simplicity of the neural networks to solve the MIL problem. Subsequently, more neural network-based MIL architectures are proposed for different applications [16, 17, 18]. And different from the instance probabilities calculation first way, Wang et al. [19] propose a novel framework that obtains the bag embedding via instance-level MIL pooling first and a classifier is built based on it to get the bag probability. Their work supplies a new approach for developing the multi-instance neural networks.

Moreover, with respect to the ability in capturing relations among instances, and between instances and bags, the attention mechanism has shown some performance advantages. It has been widely used in image and text analysis [20, 21] for now with two sub-categories: (i) task-supervised attention, and (ii) self-supervised attention. The former captures the relations between the source and target [22], and the latter computes the intra-relationship of the source [23]. Both sub-categories are essential for the MIL.

*Corresponding Author

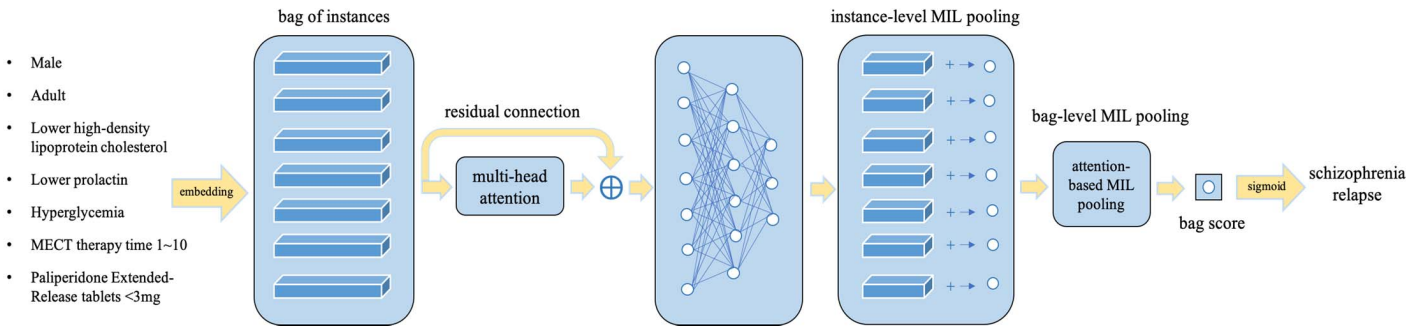

Figure 1: The overview of AMI-Net

In this study, we integrate them into a multi-instance neural network. We approach the main tasks of medical diagnosis from incomplete and low-quality data as follows: (i) mapping input instances in an embedding space [24], and instances correlating to each other over different embedding dimensions, representing some body condition, (ii) capturing the instance correlations in different embedding subspaces, (iii) learning the bag embedding, and (iv) selecting informative instances via attention mechanism to obtain the bag score. All modules are parameterized through the MIL neural network, which makes the architecture flexible and simple. This approach doesn't require data collection on purpose or data screening manually, but deal with them automatically, capturing the most useful information among a large amount of low-quality data to support the final medical diagnosis.

## II. METHODOLOGY

The overall architecture consists of an embedding layer, a multi-head attention transformer with the residual connection [25], a set of instance-wise fully connected layers, an instance-level MIL pooling layer and a bag-level MIL pooling layer followed by a sigmoid function. The overview of AMI-Net is shown in Figure 1.

### A. Multi-instance Learning

Common supervised learning aims to learn a function, mapping the input dataset X to the label set Y, in which, each object $x_i \in$ X is represented by an instance and labeled by a category or a class $Y_i \in Y$.

In MIL, the task is to learn a function mapping the input dataset $\{X_1, X_2, \ldots, X_m\}$ to the corresponding label set $\{Y_1, Y_2, \ldots, Y_m\}$ where $Y_i \in \{0, 1\}$. $X_i$ is a bag with a set of instances $\{x_{i1}, x_{i2}, \ldots, x_{i,n_i}\}$ and $n_i$ denotes to the number of instances in $X_i$. When predicting, for each bag $X_i$, if there is at least one instance labeled positive, then the bag labeled positive and otherwise is negative. The assumption can be formulated as follows:

$$Y_i = \begin{cases} 0, & \text{all } y_{ij} = 0 \\ 1, & \text{otherwise} \end{cases} \quad (1)$$

where $y_{ij}$ is the label of the $j^{th}$ instance in the $i^{th}$ bag.

The assumption above implies the underlying basis of MIL, permutation invariance property, and any permutation invariant symmetric function for solving the MIL problem can be denoted as the following function [26, 27]:

$$f(X) = \theta(\eta_{x \in X} \varphi(x)) \quad (2)$$

where $\varphi$ and $\theta$ are suitable transformations. $\eta$ is the permutation invariance function, that is well known as MIL pooling, and $\theta$ is the scoring function for a bag of instances.

About the different choices of choosing suitable $\varphi, \eta$ and $\theta$, there are two main MIL approaches:

(i) *Instance-level MIL pooling approach:* $\varphi$ is the instance transformer and the MIL pooling function $\eta$ is adopted on each instance to obtain the bag embedding for the further procession by a bag classifier $\theta$.

(ii) *Bag-level MIL pooling approach:* $\varphi$ is a transformation to return the instance scores, that are further processed by the MIL pooling $\eta$ to obtain the bag score and $\theta$ is an injective function.

**MIL with Neural Networks** Since the MIL underlying function above leaves flexibility that we can model any transformation and score function only if they follow the permutation-invariant property. Therefore, we parameterize a class of transformations through the neural network. Let $X$ be a bag of $M$ instances, the transformer $\varphi_\tau$, where $\tau$ are parameters, transforms instances to the embedding space with $K$ dimensions, that is $v_{m,K} = \varphi_\tau(x_m)$, $m \in M$. Then the bag probability of $x_m$ is determined by the transformation $\theta_\omega : \eta_{\phi_{k \in K}}(v_{m,k}) \to [0, 1]$. If using the bag-level MIL pooling approach, $\theta_\omega$ is an injective function or otherwise is parameterized by the neural networks with parameters $\omega$, and if the trainable MIL pooling methods are utilized, $\phi$ are also parameters.

**MIL pooling** As shown above, MIL pooing $\eta$ is the key step for bridging instances to bags, and different applications have their own preference for choosing MIL pooling methods. The only restriction of them on neural networks is differentiable. In MI-Net [19], max pooling, mean pooling, and log-sum-exp pooling are adopted on each instance as the instance-level MIL pooling approach and Yan et. al [18] proposed a novel dynamic pooling method integrating both instance-level and bag-level approaches.

In our proposed method, we also adopt them all on the neural network. Inspired from the sentence representation way in the

document classification problem [28], we use the sum pooling as the instance-level MIL pooling method:

$$\forall_{m=1,2,...,M}: v_m = \sum_{k=1}^{K} v_{m,k} \quad (3)$$

where $M$ and $K$ denotes the bag containing instances, and the embedding dimensions. Moreover, we propose to use attention-based MIL pooling on the bag-level to obtain the bag score, which is further mapped to the bag probability through sigmoid function.

### B. Attention-based MIL Pooling

The attention-based MIL pooling aims to assign weights, trained by the neural network, over instances. In our proposed method, it is employed in the bag-level, which is formulated as follows:

$$v = \sum_{m=1}^{M} a_m v_m \quad (4)$$

where:

$$S = W_1^T(tanh(v_m W_2) \odot sigmoid(v_m W_3))) \quad (5)$$

$$a_m = softmax(S) \quad (6)$$

where $W_1 \in \mathbb{R}^{d_{model} \times 1}$ and $W_2, W_3 \in \mathbb{R}^{d_{model} \times d_l}$ are parameters, and $\odot$ is the element-wise multiplication.

Since the $tanh$ function lacks the ability to learn the complex relations and limits the expression of non-linearity, a $sigmoid$ based function is element-wise multiplied after, which is also known as the gated mechanism [29].

Attention mechanism allows to supervise the neural network to pay more attention on the instances which are most likely to be labeled as positive [30]. It makes the model interpretable and able to detect key information from a large amount of dirty data, which is consistent to medical diagnosis process in the real life.

### C. Multi-head Attention

In our method, we propose to integrate the multi-head attention transformer [23] on the MIL neural network, to capture the intra-relationship of instances in different embedding subspaces, that perfectly fits for the medical domain since symptoms are often related to each other in different body parts or organs, and each one can be seen as a subspace. Also, standard expressions of symptoms and non-standard ones can also be linked via multi-head attention, improving model robustness to low-quality data.

The transformer takes *query* and a set of *key-value* pairs as input and output the weighted sum of *values*. The weights of *values* are calculated through the *query* and the corresponding *key* with the cosine similarity-based function. In our method, we mainly focus on the exploration of correlations among instances, therefore, *query, key* and *value* are all instances themselves. In practice, it consists of two computational parts: scaled dot-product attention and multi-head attention transformation. The architecture of the transformer is depicted as Figure 2.

**Scaled dot-product attention** Cosine similarity is computed in the subspace first for instances themselves. $Softmax$ function is used after for obtaining the final weights vector, representing the similarities and correlations of instances. Considering that the large value of instance dimensions $d_i$ likely makes the $softmax$ function to have the extremely small gradients, $\frac{1}{\sqrt{d_i}}$ is taken as the scaling factor. The final output is computed as follows:

$$Similarity(b,c) = \frac{b \cdot c}{\|b\|\|c\|} = bc^T \quad (7)$$

$$Att(X,X,X) = softmax\left(\frac{Similarity(X,X)}{\sqrt{d_i}}\right)X \quad (8)$$

where $\cdot$ is dot-product function and $X$ denotes a bag of instances.

**Multi-head transformation** It splits instance dimensions into a number of subspaces and performs scaled dot-product attention on each one in parallel, capturing the instance correlations in different subspaces. The results are concatenated together at last as the final output. Linear transformations are alternately applied in the middle. The whole process is formulated as follows:

$$MultiHead(X,X,X) = Concat(head_1, ..., head_n)W^m \quad (9)$$

$$head_i = Att(XW_i^1, XW_i^2, XW_i^3) \quad (10)$$

where $W_i^1, W_i^2, W_i^3 \in \mathbb{R}^{d_{model} \times d_k}$, $W^m \in \mathbb{R}^{hd_k \times d_{model}}$, $h$ denotes the number of heads and $head_i$ denotes the $i^{th}$ subspace.

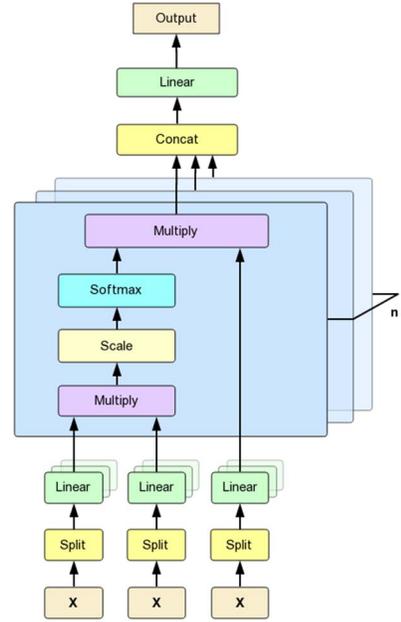

Figure 2: The architecture of multi-head attention

### III. RELATED WORK

#### A. MIL Pooling

Recently, various MIL pooling methods on the neural networks have been proposed which can be divided into two categories: non-trainable and trainable. Non-trainable methods are some simple operators such as max [19] and mean [31]. On one hand, since they are not trainable, the ability of informative instances selection is limited, but on the other hand, this way keeps the model simple and flexible, and makes sure the gradient doesn't vanish. On the contrary, the other methods are trainable ones, which are more effective and sufficient to detect the key instance, such as attention-based MIL pooling, gated attention-based MIL pooling [4] and adaptive pooling [32]. They highly enhance the performance and applicability of MIL models.

TABLE I.  EXAMPLES OF TCM AND WM DATASETS

| Dataset | Features | Diagnosis |
|---|---|---|
| TCM | Urine color yellow, Sweat, Pruritus, Coldness of extremities, Perspiration | Meridian obstruction |
| TCM | Dark red tongue, Palpitation | Not meridian obstruction |
| WM | Personal income 3000~5000, Unmarried, LOS<10 days, MECT<=1, Onset age<17, Total course<1095 days, Lorazepam tablets=0.5mg | Schizophrenia relapse |
| WM | Personal income 1000~3000, Married, High levels of prolactin, hyperglycemia, High levels of corticotrophin, LOS 25~49 days, MECT 1~10, Onset age 1-10, Risperidone<=1mg, Total course 1095~5840 days, Haloperidol injection 5mg | Not relapse |

Here, we propose to use both of them as instance-level and bag-level MIL pooling respectively to keep the model simple but efficient when capturing the valid information.

*B. Self-Attention Mechanism*

Self-attention mechanism is first proposed by Vaswani et. al [23] to solve the long-distance dependency problem and capture the relations among words in different subspaces for the natural machine translation. Their work has proven the efficiency and effectiveness of self-attention mechanism to capture syntactic and semantic information among words in text. Follow this line of research, that Shen et al. [33] use self-attention mechanism for language understanding and Tan et al. [34] adopt it for the semantic role labeling (SRL) task. Additionally, Verga et. al [35] expand this idea further to the task of biological relationship extraction and their proposed method performs noticeable well.

Motivated by the mechanism of self-attention for capturing intra-relations among words in text, we consider each patient record as a sentence with unordered words, i.e., symptoms, to explore the intra-relations of symptoms and link their standard and non-standard expressions.

## IV. EXPERIMENTS

In the experiments, we evaluated the proposed method AMI-Net on two real-world medical datasets, suitable for our approach to be applied, one from the Traditional Chinese Medicine (TCM) domain and the other from the Western Medicine (WM) domain, for the diagnosis purpose. The examples of two datasets are shown in Table 1.

*A. Datasets*

**Traditional Chinese Medicine (TCM):** The TCM dataset is collected from diabetic patients' records in a Chinese Medical Hospital in Beijing, which has been analyzed for capturing vital herb-herb interactions [36] and symptom-herb patterns [37]. In the dataset, there are 1617 outpatient records with 186 different symptoms. From patient to patient, the number of symptoms is various in 1-17. Also, the expressions of symptoms are not standardized and consistent, such as the sweat and perspiration existing in the same patient record.

The binary classification task is whether the patient has the meridian obstruction, a syndrome of TCM. Among all patients, there are 1436 labeled as negative and 181 labeled as positive. So, the dataset is highly imbalanced with the positive rate 0.112. Most importantly, there are a large number of missing values in the dataset, since it is difficult for clinicians to complete patient examinations due to the lack of patient's compliance and non-standardization of TCM information collection.

**Western Medicine (WM):** The WM dataset is provided by an AI company specialized on the medical real-world study, which collects 3927 inpatient records of schizophrenic patients, who have taken the modified electro-convulsive therapy (MECT) and improved the condition on discharge. The model aims to diagnose whether the schizophrenia relapse in three months based on the 88 physical and clinical features, such as married, unemployed, high levels of prolactin, MECT in 1-10 times and 5mg haloperidol injection. For each patient, there are at most 21 features existing and 5 at least, representing the individual patient condition. The dataset is also extremely imbalanced, that the positive rate of labels is only 0.057.

*B. Experimental Setup*

For both datasets, we padded each input record to the maximum size and the number of embedding dimensions was 128, close to the number of human organs [38]. The number of heads in the following multi-head attention transformer was set at 4. About the instance-wise fully connected layers, hidden sizes were 64 and 32 respectively. Cross-entropy was used for the final loss calculation and Adam optimizer [39] was adopted to minimize it over the training data. In relation to the hyper-parameters of Adam, we set the learning rate at 0.01, momentum parameters $\beta_1$ at 0.9, $\beta_2$ at 0.98 and $\varepsilon$ at $1e^{-8}$. In order to compare the performance, we set the binary threshold as 0.5, using AUC, Accuracy, Precision, Recall and F1-score as evaluation metrics. During the training process, the number of epochs was set at 1000 and early stopping was utilized for the best model selection according to the F1 score over the validation dataset. For the fair comparison, all experiments were run in 10-fold cross validation with five repetitions.

*C. Baselines*

The task is not only considered as a MIL problem, but also a traditional binary classification problem on the dataset in the one-hot format, that is, in the predefined feature space, learn a transformation $g: X \to [0,1]$, where $X = \{(\lambda_i, o_i)\}_{i=1}^{|X|}$ is a set of (feature, value) pairs [40] and values are all binary. If the value is missing, in the most common way, 0 is imputed to represent

the unknown condition. The following models were used as the baseline comparisons to our proposed method. The first four were built on the datasets transformed in the one-hot format.

- *Logistic Regression (LR)* [41]: A classic linear model which has been widely applied for binary classification, risk factors selection and the risk assessment scales development [42].
- *SVM* [43]: Nonlinearly mapping the input space to be high-dimensional and constructing the hyperplane set for the regression and classification tasks.
- *Random Forest* [44] and *XGBoost* [45]: Classic decision tree based algorithms, which solve the classification and regression tasks through bagging and boosting methods respectively. Both of them have gained wide attention in the medical domain for their interpretability, efficient training speed and excellent performance.
- *mi-Net* [19]: A MIL neural network using the bag-level MIL pooling approach with the max operator.
- *MI-Net, MI-Net with DS* and *MI-Net with RC* [19]: They are all proposed by Wang et al. using the instance-level MIL pooling approaches, which have achieved state-of-art performance on several classic MIL datasets.
- *Att. Net* and *Gated Att. Net* [4]: Two recent state-of-art MIL neural networks, utilizing the attention based MIL pooling on instance level to capture the relations of instance attributes.

Also, hyper-parameters of all baseline models were tuned on the validation dataset in terms of the F1-score.

## V. RESULTS AND ANALYSIS

### A. Comparison with Different Models

The results on two medical datasets are shown in the Table II and III. Our method, AMI-Net, had the best performance in terms of Precision and F1-score with the WM dataset (Table II). In fact, the F1-score from our method outperformed all other models by a large margin, although our scores on the AUC and Accuracy were slightly lower than the four classical machine-learning algorithms. Regarding the TCM dataset, which was highly imbalanced, our method outperformed all other models on Precision, Recall and F1-score.

The two datasets were incomplete and low-quality. No positive sample could be found sometimes, but our method was shown to be more robust and reliable than other models in this kind of situation. In addition, in terms of Precision, Recall and F1-score, multi-instance neural networks performed efficiently and effectively in capturing the vital information from positive samples, demonstrating the superiority of MIL in the real-life applications, with the medical domain in particular.

### B. Comparison of Different Number of Heads

In order to verify how different number of heads in multi-head attention transformer could influence the performance of AMI-net, experiments with 0, 2, 4, 8, 16 and 32 heads on both the TCM and WM datasets were set up, where 0 denoted the model without multi-head attention, using F1-score for evaluation.

TABLE II. PERFORMANCE ON THE WM DATASET

| Models | AUC | Accuracy | Precision | Recall | F1 |
|---|---|---|---|---|---|
| LR | 0.760 | 0.944 | 0.200 | 0.017 | 0.031 |
| SVM | 0.657 | **0.946** | 0 | 0 | 0 |
| Random Forest | **0.767** | 0.946 | 0 | 0 | 0 |
| XGBoost | 0.706 | 0.945 | 0.100 | 0.007 | 0.013 |
| mi-Net | 0.565 | 0.624 | 0.088 | **0.469** | 0.125 |
| MI-Net | 0.545 | 0.787 | 0.154 | 0.251 | 0.116 |
| MI-Net+DS | 0.510 | 0.621 | 0.045 | 0.383 | 0.064 |
| MI-Net+RC | 0.588 | 0.867 | 0.313 | 0.228 | 0.164 |
| Att. Net | 0.608 | 0.849 | 0.342 | 0.143 | 0.074 |
| Gated Att. Net | 0.576 | 0.832 | 0.248 | 0.140 | 0.060 |
| AMI-Net | 0.702 | 0.907 | **0.356** | 0.283 | **0.264** |

TABLE III. PERFORMANCE ON THE TCM DATASET

| Models | AUC | Accuracy | Precision | Recall | F1 |
|---|---|---|---|---|---|
| LR | **0.755** | 0.882 | 0.396 | 0.116 | 0.173 |
| SVM | 0.703 | **0.889** | 0 | 0 | 0 |
| Random Forest | 0.737 | 0.889 | 0 | 0 | 0 |
| XGBoost | 0.729 | 0.886 | 0.327 | 0.063 | 0.104 |
| mi-Net | 0.597 | 0.641 | 0.220 | 0.422 | 0.231 |
| MI-Net | 0.665 | 0.813 | 0.364 | 0.414 | 0.356 |
| MI-Net+DS | 0.586 | 0.731 | 0.358 | 0.290 | 0.179 |
| MI-Net+RC | 0596 | 0.861 | 0.353 | 0.358 | 0.312 |
| Att. Net | 0.642 | 0.861 | 0.368 | 0.244 | 0.281 |
| Gated Att. Net | 0.607 | 0.755 | 0.319 | 0.354 | 0.262 |
| AMI-Net | 0.702 | 0.818 | **0.399** | **0.468** | **0.379** |

As shown in Figure 3, when the transformer had 4 heads, our method achieved the best performance. This indicated that 4 subspaces allowed the model to have the most efficient and effective capture of intra-relations of symptoms. If it is to be interpreted using medical knowledge, it means that the condition of a patient is best considered from 4 aspects such that the symptoms are highly correlated to each other in each aspect. Coincidentally, symptoms in the TCM dataset were collected from four diagnostic methods, inspection, listening and smelling, inquiry and pulse-taking, representing four main aspects of the body condition. The experimental result was consistent with the TCM diagnosis. The results also showed that the model without the multi-head attention had the worst performance, indicating the necessity to find the correlation among symptoms, as well as linking standardized and unstandardized symptom expressions via the multi-head attention.

### C. Comparison of Different MIL Pooling Methods

In order to measure the impact of different instance-level and bag-level MIL pooling methods, the F1-scores were compared on different combinations. The max pooling [19] and the attention-based pooling [4] have previously been attempted, thus sum pooling, max pooling, attention based pooling and gated attention-based pooling, would be tested and reported in this paper.

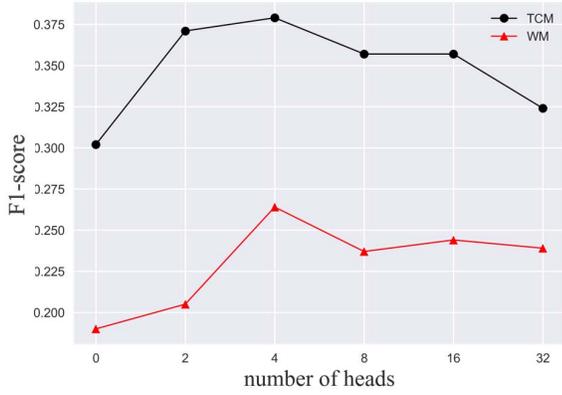

Figure 3. Comparison of different number of heads. 0 denotes the model without multi-head attention.

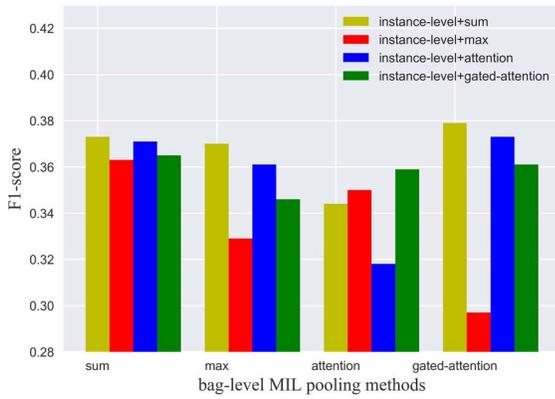

Figure 4. Comparison of different MIL pooling methods on the TCM dataset.

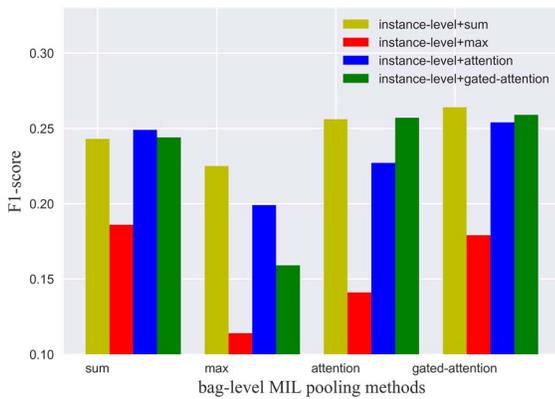

Figure 5. Comparison of different MIL pooling methods on the WM dataset.

The results were shown in Figure 4 and 5. The performance of sum pooling and gated-attention based pooling on instance-level and bag-level respectively were better than other MIL pooling methods. Moreover, the model with max pooling on instance-level was the worst, revealing that symptoms were correlated to each other on different embedding dimensions. It would be insufficient for a diagnosis if the information was only captured in one dimension,

### D. Influence of Data Noise

Since data collected from real-world studies are often in various degrees of inaccuracy and fuzziness, experiments were carried out to evaluate how the performance of the proposed model fluctuated with different data noise ratios using F1-score. There are two dimensions in data noise, namely feature noise and label noise. To test the feature noise, 1, 2, 3, 4 and 5 symptoms were randomly changed in each training sample. If the number of symptoms in a sample was lower than the required number of symptoms to change, new symptoms were added randomly. Regarding the test of label noise, the ratio of labels was inverted from 0.1 to 1.0 with the 0.1 increment in the training set.

The results of feature noise influence on different models were shown in Figure 6. Although some symptoms were randomly changed, the performance of our proposed method was still better than all other models and did not have much fluctuations. The MIL methods had more steady performance than all other machine learning algorithms, because of their ability to capture and to utilize effective information, that made them reliable in the real-life applications. The influence of label noise was displayed in Figure 7. When the proportion of inverted labels increased, the F1-score converged around 0.2 and 0.1 respectively, labelling all samples in validation set to be positive. Hence, our proposed method still performed the best in most of the time, suggesting that it is more noise resistant than others.

### E. Influence of Incomplete Data

In this section, the performance of our proposed method was studied using the incomplete datasets. 1, 2, 3, 4 and 5 symptoms were randomly deleted from each training sample. If the number of symptoms in a sample was less than the number of deletions, all symptoms would be removed. The F1-score was used for the performance comparison.

The results, illustrated in Figure 8, showed that the AMI-Net was found to be more robust than all other models under the condition of incomplete data. Since a patient does not always take all examinations and clinical measurements, a clinician very often infers missing information using his/her experience and knowledge during the diagnosis. Our proposed method offers a feasible strategy, shown to be resistant to the incomplete data, for the medical diagnosis.

### F. Visualization of Attention

The gated attention-based MIL pooling layer was able to select the most informative instances, i.e., symptoms. To make our proposed method interpretable, the visualization of attention mechanism on two examples were given in Figure 9. With the darker the color, the more important it would be. In the WM dataset, *personal income 3000-5000, unmarried, length of stay<25 days* and *MECT<=1* dominated the prediction of schizophrenia relapse. In the prediction of meridian obstruction in the TCM dataset, *decreased defecation, urine color is yellow, heavy legs* and *dropping* had larger weights to indicate their significance.

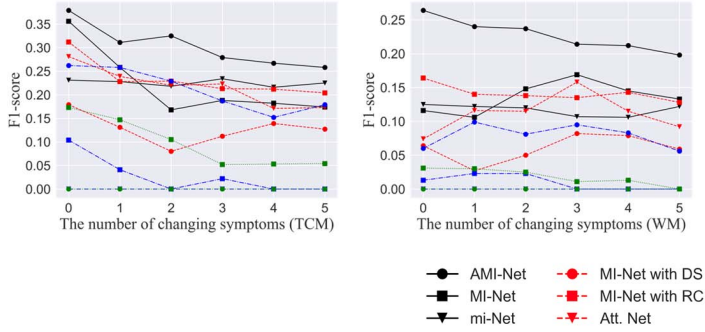

Figure 6. Test for the influence of feature noise

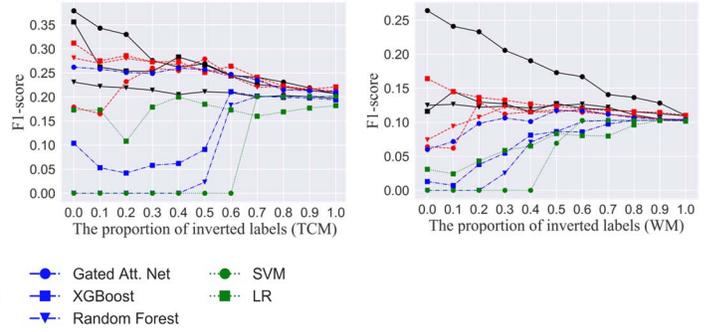

Figure 7. Test for the influence of label noise

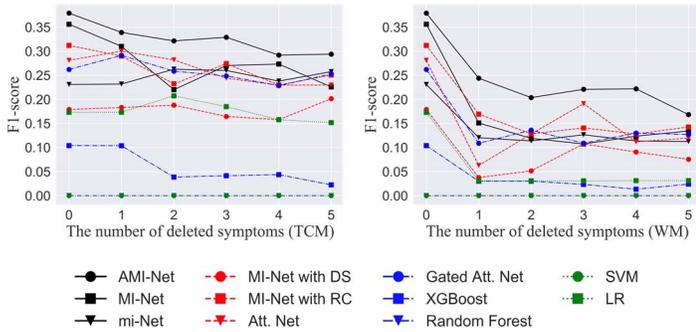

Figure 8. Test for the influence of incomplete data

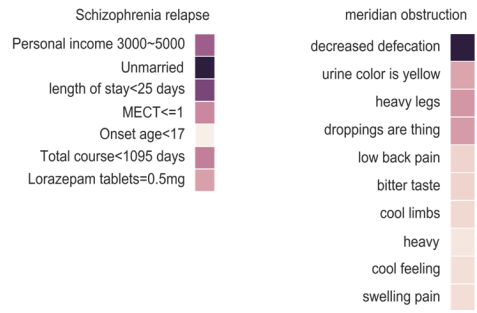

Figure 9. An example of informative instances selection

## VI. CONCLUSION

To develop an effective and efficient framework for medical diagnosis from incomplete and low-quality data, a novel attention based multi-instance neural network (AMI-Net) was proposed, that firstly captured intra-relations among instances and, secondly, selected key instances for the final classification. The experimental results demonstrated the superiority of our proposed method and MIL methods in real-life applications, in terms of Precision, Recall and F1-score. Further analysis suggested our method was interpretable, very steady and more preferable than other models under noisy and incomplete data conditions.

The study shown that AMI-Net was very effective to deal with medical diagnosis problem in real life and support the real-world study in medicine.